\documentclass[11pt]{article}

\usepackage[preprint]{acl}

\usepackage{times}
\usepackage{latexsym}

\usepackage[T1]{fontenc}
\usepackage[utf8]{inputenc}

\usepackage{microtype}

\usepackage{inconsolata}

\usepackage{graphicx}

\title{\textsc{TimeProVe}: Propose, then Verify for Efficient Long Video \\ Temporal Reasoning in Activities of Daily Living}

\author{Arkaprava Sinha, Dominick Reilly, \\ \textbf{Siddharth Krishnan, Hieu Le, Srijan Das} \\
University of North Carolina, Charlotte \\
\url{https://thearkaprava.github.io/timeprove/}}

\usepackage{graphicx}
\usepackage{algorithm}
\usepackage{algpseudocode}
\usepackage{wrapfig}
\usepackage{amsmath}
\usepackage{amssymb}
\usepackage{booktabs}
\usepackage{tabularx}
\usepackage{colortbl}
\usepackage{pifont}
\usepackage{epigraph}
\usepackage{subcaption}
\usepackage{rotating}
\usepackage{multirow}
\usepackage{soul}
\usepackage{graphicx}
\usepackage{caption}
\usepackage{xcolor} % For colored boxes
\usepackage{tikz}

\usepackage{tcolorbox}
\usepackage[normalem]{ulem} % for underline
\usepackage{calc} % For \widthof{}

\definecolor{lightblue}{RGB}{220,240,250}
\newcommand{\modelname}{\textsc{TimeProVe}}
\newcommand{\benchmark}{\textsc{OpenTSUBench}}

\newcommand{\xmark}{\ding{55}}

\begin{document}
\maketitle
\begin{abstract}
Long Video Question Answering (LVQA) requires identifying sparse, query-relevant evidence within hours-long untrimmed videos. Existing approaches either process videos densely with large vision-language models (VLMs), incurring prohibitive computational cost, or rely on sparse caption-based reasoning, which often misses temporally localized and motion-centric evidence. We introduce \textbf{\modelname}, a cost-efficient hybrid framework for temporally grounded reasoning in long videos. \modelname~first employs lightweight modules to generate action-grounded answer--evidence hypotheses and subsequently invokes an expensive VLM only for targeted verification. The core of our framework lies in the \textbf{Action-based Candidate Evidence (ACE)} module, which converts temporally localized actions into query-conditioned candidate answers and supporting evidence windows through lightweight LLM reasoning. We further introduce \textbf{\benchmark~(\textsc{otb})}, an open-ended benchmark designed to evaluate temporally grounded reasoning in real-world Activities of Daily Living (ADL) scenarios. Experiments show that \modelname~outperforms the strongest baseline on \textsc{otb} by $7.3\%$, while reducing VLM calls by $75\%$ and inference cost by $93\%$. Furthermore, without explicit temporal grounding training, \modelname~achieves competitive performance on \textsc{Charades-STA}, and reaches state-of-the-art results when enhanced with grounding VLMs. 
\end{abstract}

% Challenge/solution markers
\newcommand{\Temp}{\textbf{[T]}}   % Temporal grounding
\newcommand{\Motion}{\textbf{[M]}} % Motion/activity reasoning
\newcommand{\Eff}{\textbf{[E]}}    % Efficiency
\newcommand{\Priv}{\textbf{[P]}}   % Privacy/control

\section{Introduction}
\label{sec:intro}
In long-form Activities of Daily Living (ADL) videos, the answer to a natural-language query may hinge on a few seconds of subtle visual evidence. \textit{Taking medication, sipping water,} or \textit{picking up a small object} can be easy to miss, and visually similar activities may differ only in fine-grained hand-object interactions. For example, answering \textit{``Has the person taken their medicine, and did they drink water afterwards?''} requires finding the relevant moments in a long video and analyzing them closely enough to distinguish the intended actions. 

\begin{figure}[t]
    \centering
    \includegraphics[width=\linewidth]{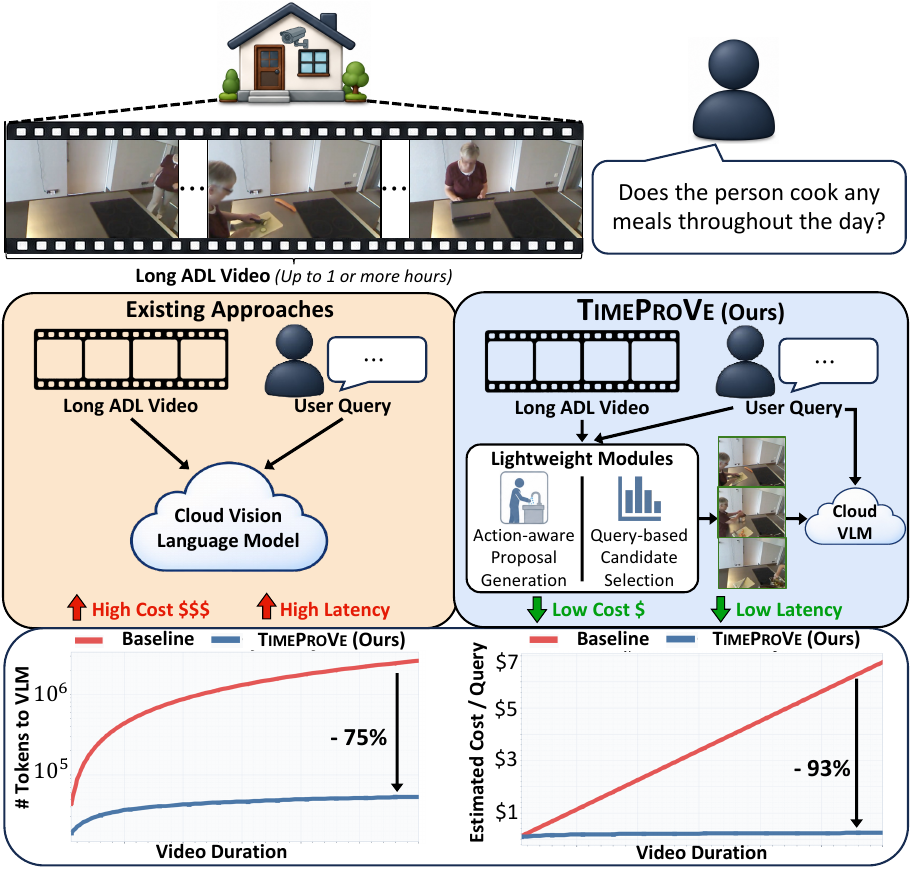}
    \vspace{-0.3cm}
    \caption{\modelname~reduces long-video LVQA cost by proposing query-relevant evidence locally before VLM verification. Instead of processing the full video, it sends only short targeted clips to the cloud VLM.}
    \vspace{-0.5cm}
    \label{fig:teaser}
\end{figure}

This makes ADL Long Video Question Answering (LVQA) fundamentally different from short, curated action recognition benchmarks~\cite{ucf, ntu120, NTU_RGB+D, smarthome, i3d}, where ${\sim}10$-second clips usually contain a single prominent action. Accurate LVQA therefore requires both temporal search and fine-grained visual-language reasoning. Large VLMs~\cite{videollama3, llavidal, cogvlm, qwen3vl, gpt4o} are effective for the latter, but applying them directly to hour-long videos is often impractical. %Visual token counts grow with video duration, which can exceed model context limits and increase latency and cost. 
For example, for a 60-minute video sampled at 1 FPS, a VLM with a SigLIP-style vision encoder~\cite{siglip} at $384\times384$ resolution with patch size $14$ produces approximately $729$ visual tokens per frame. This yields $3600 \times 729 \approx 2.6\times 10^6$ visual tokens before accounting for text prompts, timestamps, or output tokens. Thus, hour-long videos can consume more than \textit{2 million tokens}, making full-video inference difficult to scale. In practice, such inputs may exceed context limits, while cloud-based inference further introduces latency and monetary cost~\cite{gemini, chatgpt}. These constraints motivate LVQA approaches that avoid unnecessary processing of the full video while still preserving the fine-grained visual evidence needed for accurate answers.

A natural alternative to full-video VLM inference is to first caption the video~\cite{joint_modelling_captioning, ChatCaptioner, cogvlm} and perform reasoning over the resulting text, which is substantially cheaper to process than dense visual tokens~\cite{joint_modelling_captioning, ChatCaptioner, cogvlm, videoagent, videotree, videolucy, drvideo, langrepo}. However, the overall efficiency of such pipelines critically depends on the captioning model itself: while lightweight captioners can reduce computation, generating captions with large VLMs may still require repeated expensive inference over long videos. Moreover, the effectiveness of such approaches depends critically on whether the generated captions preserve the query-relevant evidence required for accurate reasoning. This is especially difficult in ADL question answering, where answers may hinge on brief, fine-grained actions rather than salient frame content. Sparse captions often omit temporal grounding, subtle motion cues, hand-object interactions, or concurrent activities. Once such evidence is absent from the text, downstream LLM reasoning cannot recover it, leading to wrong  responses.

Thus, we propose a different paradigm: keep \textit{visual} reasoning instead of converting the video to text, but restrict it to the moments where it matters. We instantiate this idea with \textbf{T}ime-aware \textbf{Pro}posal and \textbf{Ve}rification (\textbf{\modelname}), a cost-efficient hybrid framework for temporal reasoning in long videos. Rather than exhaustively processing the entire video, \modelname~first employs lightweight modules to identify answer-relevant evidence and subsequently invokes an expensive VLM only for targeted visual verification. %This design is inspired by the draft-verify paradigm in speculative decoding~\cite{long_context_spec_dec, specvlm}, where a lightweight model proposes candidates and a stronger model selectively verifies them. Unlike speculative decoding, however, the draft in our setting is not a token sequence, but an action-grounded answer--evidence hypothesis defined over time.
As illustrated in Figure~\ref{fig:teaser}, \modelname~consists of two components. The first is the \textbf{Action-based Candidate Evidence (ACE)} module, which efficiently processes the full video in a single pass. ACE employs a lightweight temporal action detector to identify and localize actions, yielding a sparse temporal action timeline describing what actions occur and when they occur. Then, conditioned on both the user query and this action timeline, a lightweight LLM generates candidate answers together with supporting evidence windows, ranked according to query relevance. These evidence windows may correspond to individual actions or short merged intervals when broader temporal context is required. The second component is a \textbf{Temporal Verifier}, which invokes an expensive VLM only on selected short RGB evidence clips. Given a candidate answer and its associated evidence window from ACE, the verifier determines whether sufficient visual evidence supports the hypothesis. If verified, \modelname~returns both the answer and its corresponding semantic and visual evidence; otherwise, it proceeds to the next candidate. Consequently, \modelname~uses VLMs for targeted verification rather than exhaustive search over long videos. In practice, \modelname~naturally supports a hybrid deployment setting, where ACE operates locally on edge devices while the Temporal Verifier leverages powerful remotely hosted VLMs.

Finally, existing LVQA benchmarks~\cite{lvbench,longvideobench} are largely restricted to multiple-choice settings, where answer options can implicitly guide evidence selection and temporal grounding is not explicitly evaluated. Therefore, we introduce \textbf{\benchmark~(\textsc{otb})}, an open-ended LVQA benchmark designed to evaluate temporally grounded reasoning in real-world ADL scenarios. \textsc{otb} requires understanding across both short atomic actions and long-horizon composite activities, making it suitable for assessing open-ended reasoning in unstructured home environments. Empirically, \modelname~outperforms the strongest baseline on \textsc{otb} by $7.3\%$ while reducing VLM invocations by $75\%$ and lowering inference cost by $93\%$. %Notably, \modelname~can operate entirely on commodity hardware, demonstrating that temporally grounded long-video understanding need not rely on continuous high-cost compute. 
We further evaluate the generality of our framework on a temporal grounding task using \textsc{Charades-STA} dataset. Despite not being explicitly trained for temporal grounding, \modelname~achieves performance comparable to specialized temporal grounding VLMs. Moreover, when temporal grounding VLMs are integrated within the ACE module, \modelname~achieves state-of-the-art performance, demonstrating the robustness of our framework.
Our key contributions are summarized below:
\vspace{-0.3cm}
\begin{itemize}
    \item We introduce \textbf{\modelname}, a novel hybrid framework that performs lightweight long-video temporal reasoning to generate action-grounded hypotheses and verifies only sparse RGB evidence using an expensive VLM.
    
    \item We design the \textbf{Action-based Candidate Evidence (ACE)} module, the first module of its kind to transform detected actions into query-conditioned answer-evidence candidates through lightweight LLM reasoning and structured reranking.
    
    \item We introduce \textbf{\benchmark}, an open-ended benchmark for temporally grounded LVQA in real-world untrimmed ADL videos.
    
    \item \modelname~achieves a $7.3\%$ improvement over the strongest baseline on \textsc{otb} while requiring substantially fewer VLM invocations and lower inference cost. Additionally, \modelname~achieves state-of-the-art performance on temporal grounding for \textsc{Charades-STA}.
\end{itemize}
\section{Related Works}
\label{sec:relatedworks}

\noindent\textbf{Vision Language Models for Long Video Understanding.}
Long Video VLMs face a token bottleneck, where na\"ively, encoding untrimmed videos produces sequences that exceed the context window of any language model and dilute relevant evidence with irrelevant background. Existing approaches for VLMs can be categorized into three main families. First, token compression methods such as LongVLM~\cite{longvlm}, VideoChat-Flash~\cite{videochatflash}, Bimba~\cite{bimba}, and STORM~\cite{jiang2025storm} hierarchically merge tokens within a fixed budget. Second, frame and token selection methods recast the problem as retrieval by ranking frames by query similarity~\cite{keyvideollm, qframe}, combinatorial coverage~\cite{framevoyager} or adaptive policies that decide how many frames to keep~\cite{flexiblefs, aks}. A third category builds memory mechanisms~\cite{moviechat, rewind} by maintaining a bounded representation that updates incrementally. However, in these methods the cost of inference scales with video length. Furthermore, token compression discards rare evidence or fine-grained details, token selection is brittle to early selection errors and may fail to capture complex temporal dependencies across multiple distant frames. Memory based representations are sensitive to memory update strategies and are prone to information drift. TimeChat~\cite{timechat}, TimeSuite~\cite{timesuite} introduce temporal grounding in VLMs by training time-aware encoders. However, these methods depend on dense timestamp-aligned instruction tuning datasets for training which is expensive for long videos. Time-R1~\cite{timer1} and Time-Zero~\cite{timezero} attempt to relax this by training with timestamp aware rewards, but Reinforcement Learning on long videos is unstable and sample-inefficient. Complementary to these approaches, \modelname~routes the dense computation to a lightweight local temporal action detection module and reserves the VLM for verification of short, query-relevant clips drawn from a sparse action prior. 

\noindent\textbf{Agentic Frameworks for Long Video Understanding.}
Recently, agent-based frameworks for long video understanding decouple the problem into perception and planning where an external LLM iteratively queries a VLM and accumulates evidence until an answer is obtained. VideoAgent~\cite{videoagent}, LangRepo~\cite{langrepo}, VideoLucy~\cite{videolucy} maintain a semantic store of captions and use an LLM to extract the answer from the captions. VideoTree~\cite{videotree} builds a hierarchical query-adaptive tree of candidate moments. More recent recursive grounding approaches such as RevisionLLM~\cite{revisionllm} and AIR~\cite{AIR} progressively refine temporal boundaries through reason-guided iteration. However, in these systems the choice of which frames to inspect is driven by an LLM operating on sparse captions or similarity scores, hence it is detached from any learned visual prior over which actions occur. Furthermore, the feedback signal between iterations is unstructured chain-of-thought rather than a grounded residual evidence representation. In contrast to these frameworks, \modelname~selects candidate windows from a learned action prior produced by an action detector. Additionally, the feedback loop is structured by an explicit residual signal to the proposal generator for calibrating windows. Unlike prior agentic systems, \modelname~returns both semantic and visual evidence to expose a provenance chain.
\begin{figure*}[ht]
    \centering
    \includegraphics[width=\textwidth,height=7.7cm,keepaspectratio]{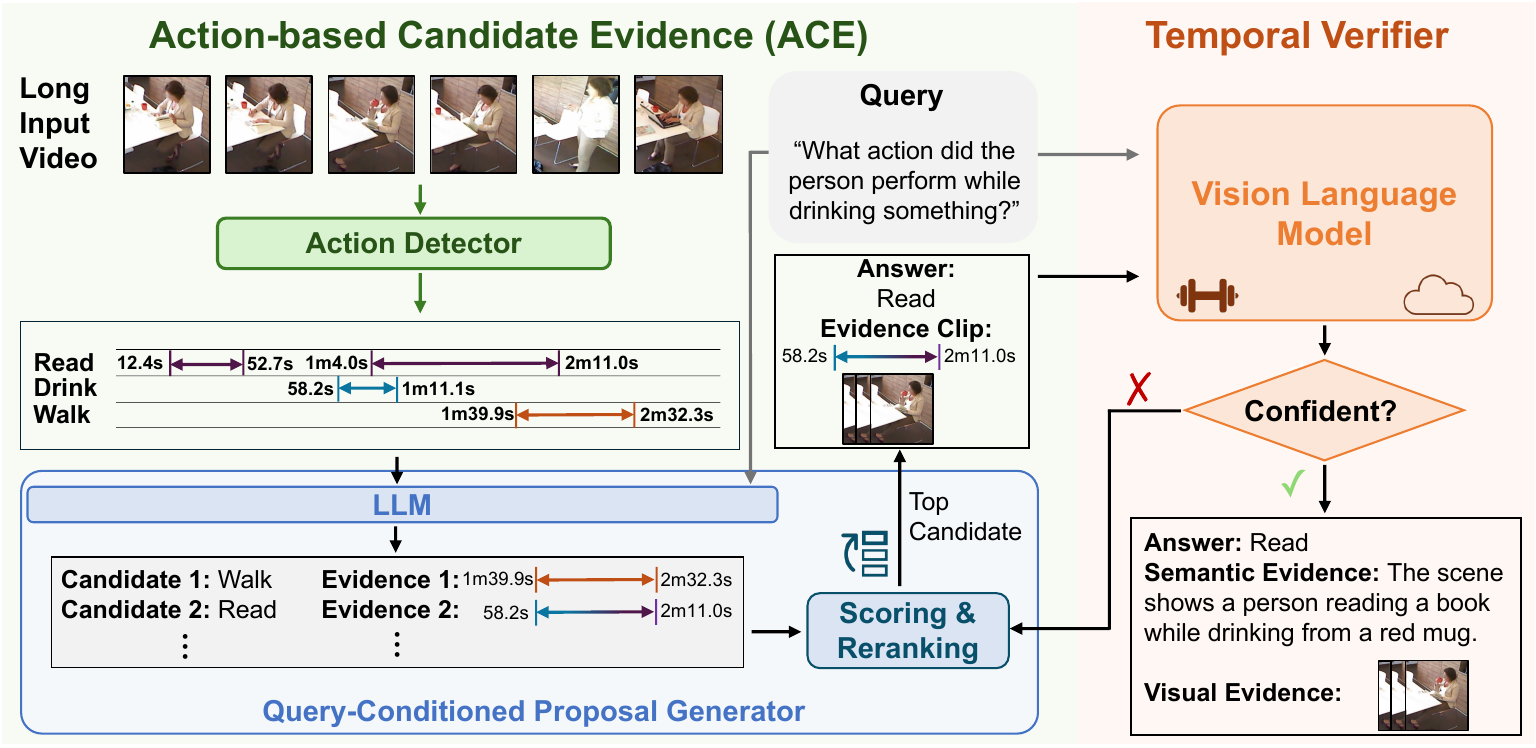}
    \vspace{-0.4cm}
    \caption{\textbf{Overview of \modelname.} ACE constructs a temporal action structure from the full video and proposes query-conditioned hypotheses. Only the selected short evidence clip is sent to the VLM for verification, avoiding full-video VLM inference.}
    \label{fig:architecture}
    \vspace{-0.5cm}
\end{figure*}

\vspace{-0.25cm}
\section{Method}
\label{sec:method}
\vspace{-0.25cm}

In long video question answering (LVQA), the goal is to answer a free-form natural language query $q$ over an untrimmed video $V$ of duration $L$. Let $V = \{f_t\}_{t=1}^{T}$ denote a sequence of $T$ frames. A direct approach is to use a large VLM to estimate $a^{\star} = \arg\max_{a} \, p(a \mid q, V)$.
However, applying a large VLM over an entire long video is prohibitively expensive, as the number of visual tokens scales with video duration.

Therefore, rather than processing the entire video or repeatedly invoking the VLM across many candidate temporal segments, our goal is to identify a small set of candidate windows $\mathcal{W} = \{w_k = [s_k, e_k]\}_{k=1}^{K}$, where $e_k - s_k \ll L$, and verify only the most promising evidence. Consequently, we propose the \textbf{T}ime-aware \textbf{P}roposal and \textbf{V}erification \textbf{F}ramework, \modelname, a cost-efficient framework for LVQA. As shown in Figure~\ref{fig:architecture}, \modelname{} consists of two main components: 
(i) an \textbf{Action-based Candidate Evidence (ACE) Module}, 
which operates over the full video using lightweight models to obtain query-conditioned candidate hypotheses, and (ii) a \textbf{Temporal Verifier}, which performs fine-grained verification only on short RGB clips selected by ACE. This design assigns broad temporal proposal generation to lightweight computation, reserving expensive VLM inference for targeted visual verification.

\subsection{Action-based Candidate Evidence (ACE)}
\label{sec:action_prior}

ACE serves as the lightweight component of \modelname. Its objective is to transform a long video into a compact temporal action representation that can be leveraged to generate candidate answer hypotheses before invoking an expensive VLM. ACE consists of two submodules: (i) \textbf{Action Detector}, which performs a single pass over the video to temporally localize actions, and (ii) \textbf{Query-conditioned Proposal Generator}, which employs an edge LLM to reason over the localized actions and produce query-conditioned candidate answers along with corresponding evidence windows.

\subsubsection{Action Detector}
\label{sec:action_detection}

First, we divide the given video $V$ into $T$ contiguous temporal segments. Following prior work in temporal action detection~\cite{dai2022mstct, mstemba}, each segment is encoded using a frozen visual backbone such as I3D~\cite{i3d} or CLIP~\cite{CLIP}, producing a feature sequence $v \in \mathbb{R}^{T \times D}$, where $D$ is the feature dimension.

The Action Detection Module predicts action probabilities over all temporal segments:
\begin{equation}
    \mathbf{P}=f_{\mathrm{act}}(v)\in[0,1]^{T\times |C|},
    \label{eq:action_probs}
\end{equation}
where $|C|$ denotes the number of action classes. We threshold $\mathbf{P}$ and decode maximal contiguous activations for each class into an event timeline:
\begin{equation}
    \mathcal{A}
    =
    \{(c_i,s_i,e_i)\}_{i=1}^{N},
    \qquad
    0 \leq s_i < e_i \leq L.
    \label{eq:event_timeline}
\end{equation}
Here, $c_i$ is the detected action label and $[s_i,e_i]$ is its temporal extent in the original video. This timeline provides a compact temporal structure over the video, indicating which actions occur and when they occur. Importantly, the full video is processed only once by the lightweight Action Detection Module. The original RGB frames are accessed again only after the Action Detector selects a short evidence window for cloud verification.

\subsubsection{Query-conditioned Proposal Generator}
\label{sec:proposal_generator}

The event timeline $\mathcal{A}$ provides a compact description of which actions occur and when they occur. 
However, it is not directly tied to the user query. 
The role of the Proposal Generator is therefore to convert this local action structure into a small set of query-conditioned hypotheses that can later be verified by the VLM.

Given the query $q$ and the event timeline $\mathcal{A}$, an edge LLM proposes candidate answers together with supporting temporal windows:
\begin{equation}
    \mathcal{H}_q=\{(a_i,w_i)\}_{i=1}^{M},
\end{equation}
where $a_i$ denotes a candidate answer and $w_i=[s_i,e_i]$ denotes the corresponding evidence window. 
The Proposal Generator constructs these evidence windows at two levels of temporal granularity.

First, for every detected event $(c_i,s_i,e_i)\in\mathcal{A}$, we create an atomic evidence window, $w_i^{\mathrm{atom}}=[s_i,e_i].$
The set of all atomic windows is denoted by $\mathcal{W}_{\mathrm{atom}}=\{w_i^{\mathrm{atom}}\}_{i=1}^{N}.$
Atomic windows preserve the finest temporal resolution of the Action Detector and are useful when the query can be answered from a single localized action.
However, many LVQA queries require context beyond a single detected event. 
For example, questions involving concurrent actions, object interactions, or temporal relations may require observing multiple neighboring or overlapping actions together. 
In order to capture such cases, the edge LLM is prompted with the query $q$ and the event timeline $\mathcal{A}$ to identify groups of atomic windows that should be considered jointly.

Let $\mathcal{G}_q=\{G_j\}_{j=1}^{J}$ denote the groups proposed by the edge LLM, where each group $G_j\subseteq\{1,\dots,N\}$ contains the indices of events that are relevant to the query. 
For each group $G_j$, we define a merged evidence window: 
\begin{equation}
    w_j^{\mathrm{merge}}
    =
    \left[
    \min_{i\in G_j} s_i,\,
    \max_{i\in G_j} e_i
    \right].
    \label{eq:merged_windows}
\end{equation}
The set of all merged windows is denoted by $\mathcal{W}_{\mathrm{merge}}=\{w_j^{\mathrm{merge}}\}_{j=1}^{J}.$
$\mathcal{W}_{\mathrm{merge}}$ allows the Proposal Generator to adapt the temporal extent of the evidence to the query. For instance, for the question \textit{``What action did the person perform while drinking something?''}, the LLM may merge the window corresponding to \textit{drink} with nearby or overlapping actions such as \textit{read}, producing an evidence window that contains the necessary context for verification.

The final query-conditioned candidate window set is then $\mathcal{W}_q=\mathcal{W}_{\mathrm{atom}}\cup\mathcal{W}_{\mathrm{merge}}.$
Each window in $\mathcal{W}_q$ is paired with a candidate answer proposed by the edge LLM to form the hypothesis set $\mathcal{H}_q$. 
This design preserves precise localization through atomic windows while allowing the evidence window to expand only when the query requires broader temporal context.

\noindent\textbf{Scoring and Reranking.}
The hypothesis set $\mathcal{H}_q$ contains candidate answers paired with temporal evidence windows. 
Although several hypotheses may be plausible, sending each corresponding RGB clip to the cloud VLM would be inefficient. 
Therefore, before any visual evidence is transmitted, the Query-conditioned Proposal Generator performs a local ranking step that estimates how likely each window is to contain the evidence needed to answer the query.

Let $\mathcal{A}(w)$ denote the detected events overlapping with window $w=[s_w,e_w]$. 
For each event $a$, $\mathcal{T}(a)$ denotes the normalized content-token set derived from its action label, containing lemmatized nouns and verbs after stop-word filtering. 
We define the window-level token set as $\mathcal{T}(w)=\bigcup_{a\in\mathcal{A}(w)}\mathcal{T}(a)$, and let $\mathcal{Q}(q)$ denote the corresponding content-token set extracted from the query. 
The ranking score combines four complementary criteria: whether the window occurs at a temporally plausible position, whether it contains an action strongly related to the query, whether it covers the query concepts collectively, and whether it remains compact enough for efficient verification.

Let $\tau=\tau(q)$ denote the temporal intent inferred from the query. 
We first compute a temporal compatibility score:
\begin{equation}
\small
R_{\mathrm{tmp}}(w,q)=
\begin{cases}
1-e_w/L, & \tau=\textsc{before},\\
s_w/L, & \tau=\textsc{after},\\
1-s_w/L, & \tau=\textsc{first},\\
s_w/L, & \tau=\textsc{last},\\
(e_w-s_w)/L, & \tau\in\{\textsc{between},\textsc{state}\},\\
1/2, & \text{otherwise}.
\end{cases}
\label{eq:temporal_score}
\end{equation}
This term acts as a soft temporal prior, biasing the ranking toward windows whose positions are compatible with the query intent.

Next, the semantic relevance score rewards the strongest action-level match within the window:
\begin{equation}
\small
R_{\mathrm{sem}}(w,q)
=
\frac{1}{Z_q}
\max_{a \in \mathcal{A}(w)}
\left|
\mathcal{Q}(q) \cap \mathcal{T}(a)
\right|,
\label{eq:semantic_score}
\end{equation}
where $Z_q=\max(|\mathcal{Q}(q)|,1)$. 
This best-match form is useful for merged windows, since a highly relevant action should not be penalized simply because the window also contains surrounding context. 
In contrast, the coverage score measures how much of the query content is represented by the window as a whole:
\begin{equation}
\small
R_{\mathrm{cov}}(w,q)
=
\frac{
|\mathcal{Q}(q) \cap \mathcal{T}(w)|
}{
Z_q
}.
\label{eq:coverage_score}
\end{equation}
Thus, $R_{\mathrm{sem}}$ favors a strong local match, while $R_{\mathrm{cov}}$ favors windows that jointly cover multiple query-relevant concepts. 
Finally, we use $R_{\mathrm{len}}(w)=(e_w-s_w)/L$ to penalize unnecessarily long clips.
The final local ranking score is:
\begin{equation}
\small
\begin{aligned}
R(w \mid q)
&=
R_{\mathrm{tmp}}(w,q) 
+
R_{\mathrm{sem}}(w,q) \\
&\quad+
R_{\mathrm{cov}}(w,q)
-
R_{\mathrm{len}}(w).
\end{aligned}
\label{eq:relevance_score}
\end{equation}
Consequentially, sorting the candidates by this score yields:
\begin{equation}
\small
\mathcal{H}_q^{*}
=
\big[
(a_{(1)},w_{(1)}),
\dots,
(a_{(M)},w_{(M)})
\big],
\label{eq:ranked_candidates}
\end{equation}
where $R(w_{(j)}\mid q)\geq R(w_{(j+1)}\mid q)$ for $j=1,\dots,M-1$.

This scoring function is the main mechanism that turns the raw action timeline into a query-conditioned hypothesis structure. 
Rather than treating all detected events as equally relevant, it organizes candidate answers by their temporal, semantic, and cost-aware compatibility with the query. 
Consequently, the top-ranked hypothesis serves as the most likely answer-evidence pair and is selected for expensive VLM verification.

\subsection{Temporal Verifier}
\label{sec:temporal_verifier}

In \modelname, ACE efficiently narrows the search space, but the action timeline alone cannot capture all fine-grained visual details needed for answering, such as objects, attributes, interactions, and scene context. Therefore, \modelname~uses a cloud VLM only as a verifier over selected short clips, rather than as a full-video reasoner.

At verification step $t$, let $(a_t,w_t)$ be the highest-ranked unverified hypothesis from $\mathcal{H}_q^{*}$, where $w_t=[s_t,e_t]$. We extract the corresponding RGB evidence clip $\widetilde{V}_t=V[s_t,e_t]$ and send only this clip, together with the query and candidate answer, to the VLM, $(c_t,\hat{a}_t,d_t) = f_{\mathrm{vlm}}(\widetilde{V}_t,q,a_t).$
Here, $c_t\in\{0,1\}$ indicates whether the clip contains sufficient visual evidence, $\hat{a}_t$ is the verified answer, and $d_t$ is the semantic evidence extracted from the clip.

If $c_t=1$, \modelname~returns:
\begin{equation}
    (a^{*},\mathcal{S}^{*},\mathcal{V}^{*})
    =
    (\hat{a}_t,d_t,\widetilde{V}_t),
    \label{eq:verified_output}
\end{equation}
where $a^{*}$ is the final answer, $\mathcal{S}^{*}$ is the semantic evidence, and $\mathcal{V}^{*}$ is the visual evidence clip. If $c_t=0$, the verifier rejects the hypothesis and proceeds to the next candidate in $\mathcal{H}_q^{*}$. The process stops when a candidate is accepted or the verification budget is exhausted.
\vspace{-3mm}
\section{\benchmark~(\textsc{otb})}
\label{sec:tsubench}
LVQA is most useful when a model can not only answer a question, but also identify the temporal evidence that supports the answer. This requirement is especially important for ADL, where the relevant evidence may occupy only a few seconds within a long, visually redundant recording. Existing LVQA benchmarks often emphasize multiple-choice evaluation, report aggregate accuracy without diagnostic breakdowns, or omit precise temporal evidence. As a result, they make it difficult to evaluate whether a model is genuinely grounded or merely producing the right answer from language priors or dataset biases.

Therefore, we introduce \textbf{\benchmark} (\textsc{otb}), an open-ended, temporally grounded QA benchmark built on the Toyota Smarthome Untrimmed Dataset (TSU)~\cite{Dai_2022_PAMI}. \textsc{otb} contains $3{,}567$ question-answer pairs over $185$ untrimmed ADL videos, with an average video duration of $21$ minutes. Each question is paired with one or more supporting temporal intervals, allowing models to be evaluated both on answer correctness and on whether they localize the evidence used to answer the question.

The benchmark is constructed from timestamped TSU action annotations. We first canonicalize each video into an action timeline, instantiate a library of templated questions over the timeline, process them into natural language using a constrained LLM, and then filter them for ambiguity, triviality, and redundancy. Full construction details, prompts, filtering rules, and additional statistics are provided in the Appendix.
\begin{table*}[h!]
\centering
\caption{\textbf{Comparison with State-of-the-Art on \benchmark.}}
\vspace{-0.3cm}
\setlength{\tabcolsep}{7pt}
\scalebox{0.6}{
\begin{tabular}{l|cc|cccccc}
\toprule
\multirow{2}{*}{\textbf{Method}} & \multirow{2}{*}{\textbf{LLM}} & \multirow{2}{*}{\textbf{VLM}} & \textbf{Object} & \textbf{Temporal} & \textbf{Compositional} & \textbf{State} & \textbf{Long-Horizon} & \multirow{2}{*}{\textbf{Overall}}\\
& & & \textbf{Centric} & \textbf{Positioning} & \textbf{Actions} & \textbf{Transition} & \textbf{Sparse Evidence} & \\
\midrule
\rowcolor{black!9} 
    \multicolumn{9}{l}{\emph{SFT-Based Frameworks}} \\
VideoLLaMA3~\cite{videollama3}  & Qwen 2 & - & 7.8 & 22.3 & 21.5 & 71.9 & 15.7 & 22.6 \\
Qwen2.5-VL~\cite{qwen25} & Qwen 2.5 & - & \textbf{72.1} & 40.9 & 26.6 & 27.2  & 35.8 & 39.3 \\
VideoChat-Flash~\cite{videochatflash} & InternLM2 & - & 63.6 & 36.4 & 29.3  & 66.7 & 29.7 & 37.8 \\
VTimeLLM~\cite{vtimellm} & LLaMA-2-7B & - & 55.8 & 27.0 & 32.5 & 65.9 & 29.7 & 33.1 \\
TimeChat~\cite{timechat}  & LLaMA-2-7B & - & 62.8 & 30.7 & 13.1  & 55.7 & 21.2 & 30.4 \\
Time-R1~\cite{time-r1} & Qwen2.5VL & - & 50.8 & 35.6 & 26.9  & 49.2 & 28.8 & 34.9 \\
TimeSuite~\cite{timesuite} & Mistral-7B & - & 68.2 & 31.4 & 23.9  & 80.9 & 26.7 & 35.4 \\
\midrule
\rowcolor{black!9} 
    \multicolumn{9}{l}{\emph{Agentic Frameworks}} \\
VideoTree~\cite{videotree} & GPT-5.1 & - & 36.8 & 30.3 & 19.4 & 33.7 & 21.2 & 27.3 \\
AVP~\cite{avp}  & GPT-4o & GPT-4o & 11.5  & 20.2 & 9.4 & 42.3 & 3.9 & 14.4  \\
GPT-4o~\cite{gpt4o}  & GPT-4o & GPT-4o & 27.9 & 27.3 & 9.4 &  67.7 & 11.2 & 23.8 \\
Videoagent~\cite{videoagent}  & GPT-4o & GPT-4o & 65.1  & 35.6 & 17.5 & 34.9 & 28.6 & 33.9  \\
\midrule
\rowcolor{lightblue}
\modelname~(Ours)  & Gemma4-2B & VLMA3 & 53.9 & 33.5 & 27.7  & \textbf{82.9} & 31.7 & 37.3 \\
\rowcolor{lightblue}
\modelname~(Ours)  & Qwen2-7B & VLMA3 & 53.5 & 42.0 & 31.7  & 78.9 & \textbf{36.1} & 42.7 \\
\rowcolor{lightblue}
\modelname~(Ours)  & Gemma4-2B & GPT-4o & 49.2 & \textbf{47.9} & \textbf{37.1}  & 70.3 & 35.7 & \textbf{45.1} \\
\bottomrule
\end{tabular}}
\label{tab:sota}
\vspace{-6mm}
\end{table*}

\vspace{-3mm}
\section{System Evaluation}
\label{sec:eval}
\vspace{-2mm}
\subsection{Implementation Details}
For the Action Detector in ACE module, we use MS-Temba~\cite{mstemba}, a lightweight temporal action detector with 17M parameters. Videos are divided into contiguous 16-frame windows, and segment-level features are extracted using either CLIP-L/14~\cite{CLIP} or I3D~\cite{i3d}. MS-Temba is either trained on Toyota Smarthome Untrimmed~\cite{Dai_2022_PAMI} or Charades~\cite{charades}, depending on the downstream benchmark, and predicts class-wise action probabilities for each temporal segment. We threshold the probabilities at $\theta=0.5$ and decode contiguous activations into the event timeline used by ACE. For the Query-conditioned Proposal Generator, we use Gemma4-2B~\cite{gemma4} and Qwen2-7B~\cite{qwen} to produce candidate answer-evidence pairs from the event timeline. For the Temporal Verifier, we evaluate VideoLLaMA3~\cite{videollama3} or GPT-4o~\cite{gpt4o}. We provide the complete prompts used for the Proposal Generator and the Temporal Verifier in the Appendix.

\iffalse
In \modelname, the local Action Prior consists of an Action Detection Module and an edge LLM-based Proposal Generator. 
For the Action Detection Module, we use MS-Temba~\cite{mstemba}, a lightweight temporal action detector with 17M parameters. 
Following the temporal action detection setup, videos are divided into contiguous windows of 16 frames, and segment-level features are extracted using either CLIP-L/14~\cite{CLIP} or I3D~\cite{i3d}. 
MS-Temba is trained on the Toyota Smarthome Untrimmed dataset~\cite{Dai_2022_PAMI} and Charades~\cite{charades} and predicts class-wise action probabilities for each temporal segment. 
We threshold these probabilities with $\theta=0.5$ and decode contiguous activated segments into the event timeline used by the Action Prior. For the Proposal Generator, we use Gemma~\cite{gemma4} as the lightweight edge LLM backbone. For the cloud Temporal Verifier, we evaluate both open-source and closed-source VLMs. For the open-source setting, we use VideoLLaMA3~\cite{videollama3}, which consists of a SigLIP~\cite{siglip} vision encoder, a two-layer MLP projector with GELU activation, and a Qwen2 language model. 
For the closed-source setting, we use GPT-4o~\cite{gpt4o} through API calls. 
We provide the complete prompts used for the edge LLM Proposal Generator and the Temporal Verifier in Appendix.
\fi

\vspace{-3mm}
\subsection{Comparison with State-of-the-Art}
\vspace{-1mm}
In Table~\ref{tab:sota}, we compare \modelname~with two families of methods, supervised fine-tuned (SFT) VLMs and agentic VLM-based frameworks. 
Among SFT-based methods, VideoLLaMA3 achieves the lowest performance, while temporal grounding models such as TimeChat, VTimeLLM, Time-R1, and TimeSuite improve performance on specific categories through time-aware instruction tuning but remain limited overall.
Interestingly, \modelname~with VLMA3 as the verifier achieves stronger overall performance. Using Gemma4-2B in ACE, \modelname~achieves an improvement of $14.7\%$ over the baseline with the same VLMA3 as Temporal Verifier. With Qwen2-7B in ACE, \modelname~achieves further improvement of $5.4\%$, with gains across temporal positioning, compositional actions, and long-horizon sparse evidence.  

Among the agentic frameworks, VideoAgent performs strongly on object-centric questions, but drops on state transitions and sparse-evidence reasoning, suggesting that frame-level captioning can capture salient objects while missing temporally specific evidence. In contrast, \modelname~with GPT-4o as verifier achieves an improvement of $21.3\%$ over the GPT-4o baseline and $11.2\%$ over VideoAgent.
Furthermore, \modelname~consistently achieves a performance improvement on Temporal Positioning and Long-Horizon Sparse Evidence category, demonstrating its effectiveness in capturing temporal dependencies among actions in the long videos. 

\begin{table*}[t]
\centering
\label{tab:efficiency_ablation_combined}

\begin{minipage}[t]{0.3\textwidth}
\vspace{0pt}
\centering
\captionof{table}{\textbf{Ablating the Components of \modelname.}}
\vspace{-0.3cm}
\scalebox{0.7}{
\begin{tabular}{cccc}
\toprule
\multirow{2}{*}{\shortstack{\textbf{Action}\\\textbf{Detection}}}
& \multicolumn{2}{c}{\textbf{Proposal Generator}}
& \multirow{2}{*}{\textbf{Perf.}} \\
\cmidrule(lr){2-3}
& \textbf{Edge} & \textbf{Score} & \\
& \textbf{LLM} & \textbf{Rerank} & \\
\midrule
\xmark & \xmark & \xmark & 24.7 \\
\checkmark & \xmark & \xmark & 36.4 \\
\checkmark & \checkmark & \xmark & 40.0 \\
\checkmark & \checkmark & \checkmark & \textbf{42.7} \\
\bottomrule
\end{tabular}}
\label{tab:components}
\end{minipage}
\hspace{2mm}
\begin{minipage}[t]{0.39\textwidth}
\vspace{0pt}
\centering
\captionof{table}{
\textbf{Efficiency comparison of \modelname~with LVQA baselines.}}
\vspace{-0.3cm}
\scalebox{0.8}{
\begin{tabular}{lcccc}
\toprule
\textbf{Method} & \textbf{Acc.} & \textbf{\# Calls} & \textbf{Dur.} & \textbf{Lat.} \\
\midrule
Caption-Based & 24.7 & 16.8 & 1004.8 & 55.0 \\
Uniform Sampling & 34.7 & 16.8 & 1004.8 & 27.0 \\
Full-Video & 35.0 & 1.0 & 180.0 & 17.6 \\
Retrieval-Based & 33.9 & 7.0 & 10.0 & 35.0 \\
\modelname & 44.8 & 8.3 & 123.6 & 18.7 \\
\bottomrule
\end{tabular}}
\label{tab:efficiency}
\end{minipage}
\hfill
\begin{minipage}[t]{0.23\textwidth}
\vspace{0pt}
\centering
\scalebox{0.7}{
\includegraphics[width=\linewidth]{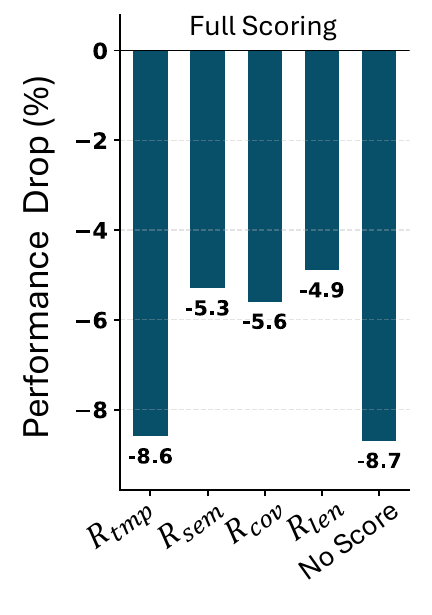}
}\vspace{-5mm}
\captionof{figure}{\textbf{Scoring Metrics}}
\label{fig:score}
\end{minipage}
\vspace{-6mm}
\end{table*}

\vspace{-3mm}
\subsection{System Diagnosis}
\vspace{-1mm}
\noindent\textbf{Effect of framework components.} Table~\ref{tab:components} ablates the main components of \modelname. Without the Action Detector, edge LLM, or scoring module, the system reduces to a weak caption-style or unguided baseline. Adding the Action Detector improves performance considerably, showing that the detected action timeline is a strong source of temporal structure. This validates the central assumption of our approach that grounding evidence in temporally localized actions is essential for temporal reasoning in untrimmed ADL videos. Adding the edge LLM further improves performance by $3.6\%$. This gain indicates that query-conditioned reasoning is necessary to identify which temporal windows should be merged to answer the query. Finally, incorporating proposal scoring and reranking achieves the best performance, indicating that heuristic-guided structuring of candidate proposals before verification is crucial for effective temporal reasoning.

\noindent\textbf{Efficiency Analysis.} 
Table~\ref{tab:efficiency} highlights the accuracy-efficiency tradeoff among LVQA methods. Caption-based and uniform-sampling methods process much longer video duration through repeated VLM calls, yet remain less accurate. In contrast, full-video inference has low latency but its comparable accuracy shows that exposing the VLM to more video frames does not necessarily yield better reasoning when evidence is sparse. Retrieval-based selection is efficient in processed duration, but its low performance suggests that generic retrieval often misses action-relevant evidence. In contrast, \modelname~achieves the best accuracy with minimal latency overhead, because ACE narrows the search before VLM verification. This shows that the key efficiency gain is not merely reducing calls or duration in isolation, but selecting clips that are likely to contain the answer. \\

\begin{table}[h]\vspace{-6mm}
\centering
\caption{\textbf{Temporal Grounding on Charades-STA}}
\vspace{-0.3cm}
\scalebox{0.6}{
\begin{tabular}{lccc}
\toprule
\textbf{Method} %& \multicolumn{3}{c}{\textbf{Charades-STA}} \\
% \cmidrule(lr){2-4}
& \textbf{R1@0.3} & \textbf{R1@0.5} & \textbf{R1@0.7} \\
\midrule
VideoChat2~\cite{mvbench} & 9.6 & 3.4 & 1.4\\
TimeChat~\cite{timechat} & - & 32.2 & 13.4\\
VTimeLLM~\cite{vtimellm} & 51.0 & 27.5 & 11.4\\
ChatVTG~\cite{chatvtg} & 52.7 & 33.0 & 15.9\\
TimeSuite~\cite{timesuite} & 69.9 & 45.3 & 24.0\\
Time-R1~\cite{timer1} & 77.7 & 61.5 & \textbf{36.8}\\
\rowcolor{lightblue}
\modelname~(+ Action Detector) & 52.1 & 27.3 & 10.7\\
\rowcolor{lightblue}
\modelname~(+ TimeSuite) & 71.2 & 50.1 & 24.6\\
\rowcolor{lightblue}
\modelname~(+ Time-R1) & \textbf{78.2} & \textbf{62.0} & 36.1\\
\bottomrule
\end{tabular}}
\label{tab:tvg}
\end{table}
\vspace{-3mm}

\noindent\textbf{Effect of scoring metrics.} 
Figure~\ref{fig:score} evaluates the drop in performance on the ablation of each term in the local ranking score. The full scoring function performs best, showing that effective evidence selection requires combining temporal, semantic, and cost-aware cues. Temporal compatibility is the most influential since removing it causes the largest drop and brings performance close to the no scoring variant. This demonstrates that for long-video temporal reasoning, it is important for the evidence window to be consistent with the temporal intent of the query. Additionally, the remaining terms contribute complementary improvements, substantiating that all components of the ranking mechanism are essential for refining the final ordering of candidate hypotheses.

\vspace{-2mm}
\subsection{Generalization to Temporal Grounding}
\vspace{-1mm}
Although \modelname~is specifically designed for open-ended LVQA, its intermediate representation, i.e., localized evidence window from ACE, is inherently temporal. This makes it natural to ask whether the same mechanism can transfer to temporal grounding, where the task is to localize the interval corresponding to a language query. We evaluate this on Charades-STA in Table~\ref{tab:tvg}. 
Complementary to LVQA, Charades-STA evaluates short, free-form temporal grounding rather than open-ended reasoning over long ADL videos. Accordingly, \modelname~with only the Action Detector is limited at stricter IoU thresholds, since action detectors provide event-level segments rather than the fine boundary alignment required by temporal grounding. Nevertheless, this variant remains competitive with several general video-language grounding models, suggesting that action timelines provide useful temporal structure even beyond the target LVQA setting. Notably, while strong baselines such as TimeSuite and Time-R1 achieve competitive performance on \textsc{Charades-STA}, they substantially underperform on \textsc{otb}, highlighting the challenges posed by temporally grounded reasoning in long ADL videos.

Moreover, \modelname~is not tied to a specific detector. When combined with stronger temporal localization backbones, it improves over TimeSuite by $1.3$, $4.8$, and $0.6$ points respectively. With Time-R1, it further improves the looser and medium IoU thresholds while remaining comparable at the strictest threshold. These gains indicate that ACE acts as a reusable query-conditioned evidence selection layer that can plug into stronger temporal grounding modules to improve evidence localization beyond answer generation.
\vspace{-3mm}
\section{Conclusion}
\label{sec:conclusion}
\vspace{-2.5mm}
We propose \modelname, a cost-efficient framework for temporally grounded LVQA. Instead of processing an entire long video with a VLM, \modelname~uses ACE to generate action-grounded answer--evidence hypotheses and invokes an expensive VLM only for targeted verification of short RGB clips. We further introduce \benchmark, an open-ended benchmark for evaluating temporal reasoning in real-world ADL videos. Experiments show that \modelname~improves accuracy while substantially reducing VLM calls and inference cost, and further generalizes to temporal grounding when combined with stronger temporal localization modules. 

\newpage
\section{Limitations}
\label{sec:limitations}
\vspace{-2mm}
\modelname~assumes that the answer-relevant evidence can be captured by a small number of localized or merged action windows. This is well suited to ADL-style reasoning, but questions requiring diffuse scene understanding over very long intervals may require broader evidence aggregation. Additionally, while \modelname~substantially reduces VLM usage, the final verification step still depends on the visual reasoning ability of the chosen VLM. Improving adaptive evidence aggregation, and verifier calibration are promising directions for future work. \vspace{-3mm}

\section*{Acknowledgements}
\vspace{-2mm}
This work was supported in part by the National Science Foundation (IIS-2245652) and the University of North Carolina at Charlotte. Computational resources were provided by the NSF National AI Research Resource Pilot (NAIRR240338) and NCShare.
\vspace{-8mm}

% Custom bibliography entries only
% \newpage
\bibliography{ref}
\newpage
\appendix

\section*{Appendix}
\section*{Overview}
The Supplementary material is organized as follows:

\begin{itemize}
    \item Section~\ref{app:noise}: \modelname's robustness to noisy Action Priors.
    \item Section~\ref{app:evidence}: Evidence Grounding Capability of \modelname
    \item Section~\ref{app:otb}: OpenTSUBench (\textsc{otb}): Construction and Evaluation Details
    \item Section~\ref{app:algo}: Algorithmic Framework of \modelname
    \item Section~\ref{app:prompts}: Prompts
    \item Section~\ref{app:qual}: Qualitative Examples
\end{itemize}

\begin{figure}[h]
  \centering
  \includegraphics[height=5cm, width=0.3\textwidth]{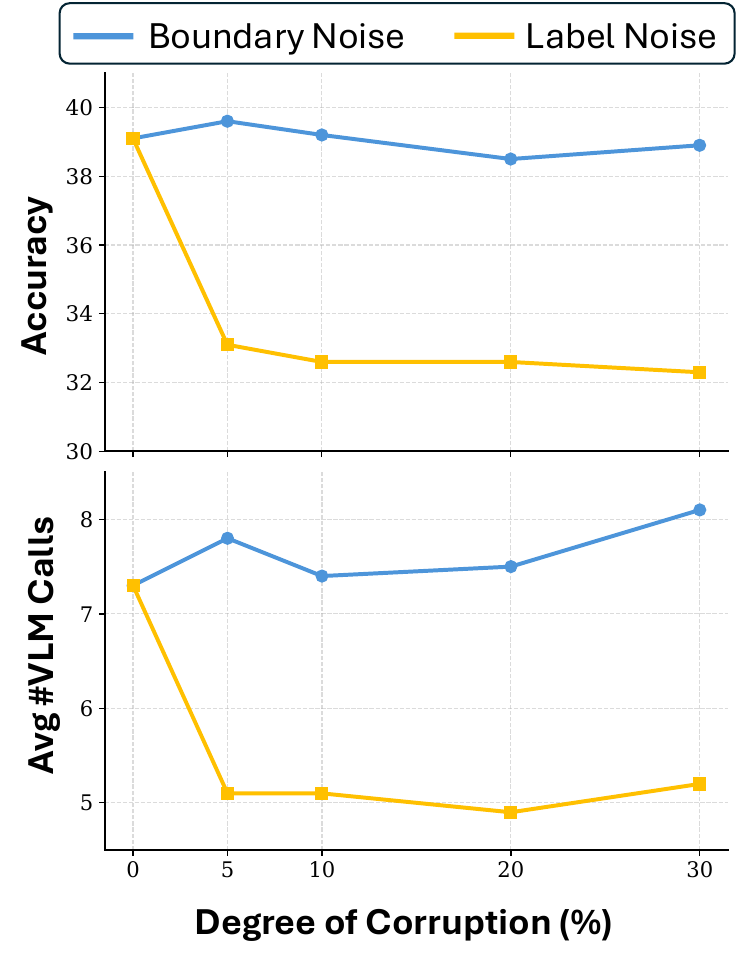}
  \caption{Effect of boundary and label noise on Accuracy and number of VLM calls.}
  \label{fig:noise}
\end{figure}

\section{\modelname's robustness to noisy Action Priors}
\label{app:noise}
Since \modelname~uses the ACE module to construct candidate evidence windows from an action timeline, it is important to understand the sensitivity of the framework to imperfect action detection. We therefore evaluate \modelname~under corrupted event timelines. Starting from the detected timeline $\mathcal{A}=\{(c_i,s_i,e_i)\}_{i=1}^{N}$, we introduce two types of perturbations at increasing noise levels. For \textit{label noise}, we randomly replace the action labels $c_i$ of a fraction $p$ of events with labels sampled from the remaining action vocabulary, while keeping their temporal boundaries fixed. For \textit{boundary noise}, we perturb each event boundary as $\tilde{s}_i=s_i+\epsilon_i^s$ and $\tilde{e}_i=e_i+\epsilon_i^e$, where $\epsilon_i^s,\epsilon_i^e\sim\mathcal{N}(0,\sigma^2)$, followed by clipping to the valid video range and enforcing $\tilde{s}_i<\tilde{e}_i$. We conduct this analysis on a randomly sampled subset of 1000 QA pairs from \textsc{otb}.

Figure~\ref{fig:noise} demonstrates \modelname's effectiveness in the absence of perfect action timeline. Boundary perturbations have only a limited impact up to moderate corruption levels, indicating that ACE can still propose windows that overlap with the relevant evidence even when temporal boundaries are imprecise. Label noise is more challenging because it directly affects query-conditioned proposal generation and reranking; nevertheless, the performance drop remains bounded. This is because the Temporal Verifier does not rely solely on action labels, it validates the selected RGB clip with the VLM before producing the final answer. These results show that the local action timeline acts as an efficient guide for evidence selection, while final prediction remains grounded in visual verification, making \modelname~robust to imperfections in the local temporal model.

\begin{table}[h]
\centering
\caption{\textbf{Evidence Grounding on \textsc{otb}}}
\scalebox{0.8}{
\begin{tabular}{ccc}
\toprule
\textbf{Temporal} & \textbf{Captioning} & \multirow{2}{*}{\textbf{\modelname}} \\
\textbf{Grounding} & \textbf{Based} \\
\midrule
NA & 9.1 & 22.3 \\
\bottomrule
\end{tabular}}
\label{tab:evidence}
\end{table}

\section{Evidence Grounding Capability of \modelname}
\label{app:evidence}

Table~\ref{tab:evidence} evaluates the quality of the temporal evidence selected by \modelname~on \textsc{otb}. Although temporal-grounding methods are trained with timestamp-based instruction-tuning data, they fail to reliably localize evidence in long, untrimmed ADL videos. Captioning-based approaches provide a stronger point of comparison: they partition the video into fixed-size clips, generate timestamped clip-level captions, and aggregate them into a video-level description, allowing each answer to be traced back to the corresponding captioned clip. We report the temporal IoU between the selected evidence and the ground-truth evidence intervals, using a 10-second buffer around the reference window. \modelname~achieves a tIoU of 22.3, more than twice that of the captioning-based baseline. This result shows that ACE identifies answer-relevant evidence more precisely than dense captioning, despite operating through lightweight action-grounded proposals rather than exhaustive caption generation.

\begin{table*}[t]
\centering
\small
\caption{Comparison of \textsc{otb} with contemporary long-video QA benchmarks. ``Temporal GT'' indicates presence of ground-truth supporting intervals. ``Open-ended'' indicates free-form answers rather than multiple choice. ``Multi-interval'' indicates support for answers that depend on temporally separated evidence.}
\label{tab:otb-vs-prior}
\scalebox{0.84}{
\begin{tabular}{lcccccc}
\toprule
Benchmark & Avg.\ len. & Open-ended & Temporal GT & Multi-interval & Stratified & Domain \\
\midrule
AGQA~\cite{agqa}   & 30 s  & \checkmark        & \ding{55}         & \ding{55}         & \ding{55}         & ADL \\
ActivityNet-QA~\cite{activitynetqa}   & 3 min  & \checkmark        & \ding{55}         & \ding{55}         & \ding{55}         & internet \\
NExT-QA~\cite{nextqa}         & 44 s   & \ding{55} (MCQ)   & weak              & \ding{55}         & \checkmark        & social   \\
Video-MME~\cite{videomme}        & 17 min & \ding{55} (MCQ)   & \ding{55}         & \ding{55}         & \checkmark        & varied   \\
MLVU~\cite{mlvu}             & 12 min & partial           & partial           & \ding{55}         & \checkmark        & varied   \\
LongVideoBench~\cite{longvideobench}   & 12 min & \ding{55} (MCQ)   & weak              & \ding{55}         & \checkmark        & varied   \\
% MoVQA            & 30 min & \ding{55} (MCQ)   & \ding{55}         & \ding{55}         & \checkmark        & movies   \\
\midrule
\textbf{\textsc{otb} (ours)} & 21 min & \checkmark & \checkmark & \checkmark & \checkmark & ADL \\
\bottomrule
\end{tabular}}
\end{table*}

\begin{figure*}[h]
  \centering
  \includegraphics[height=9cm, width=\textwidth]{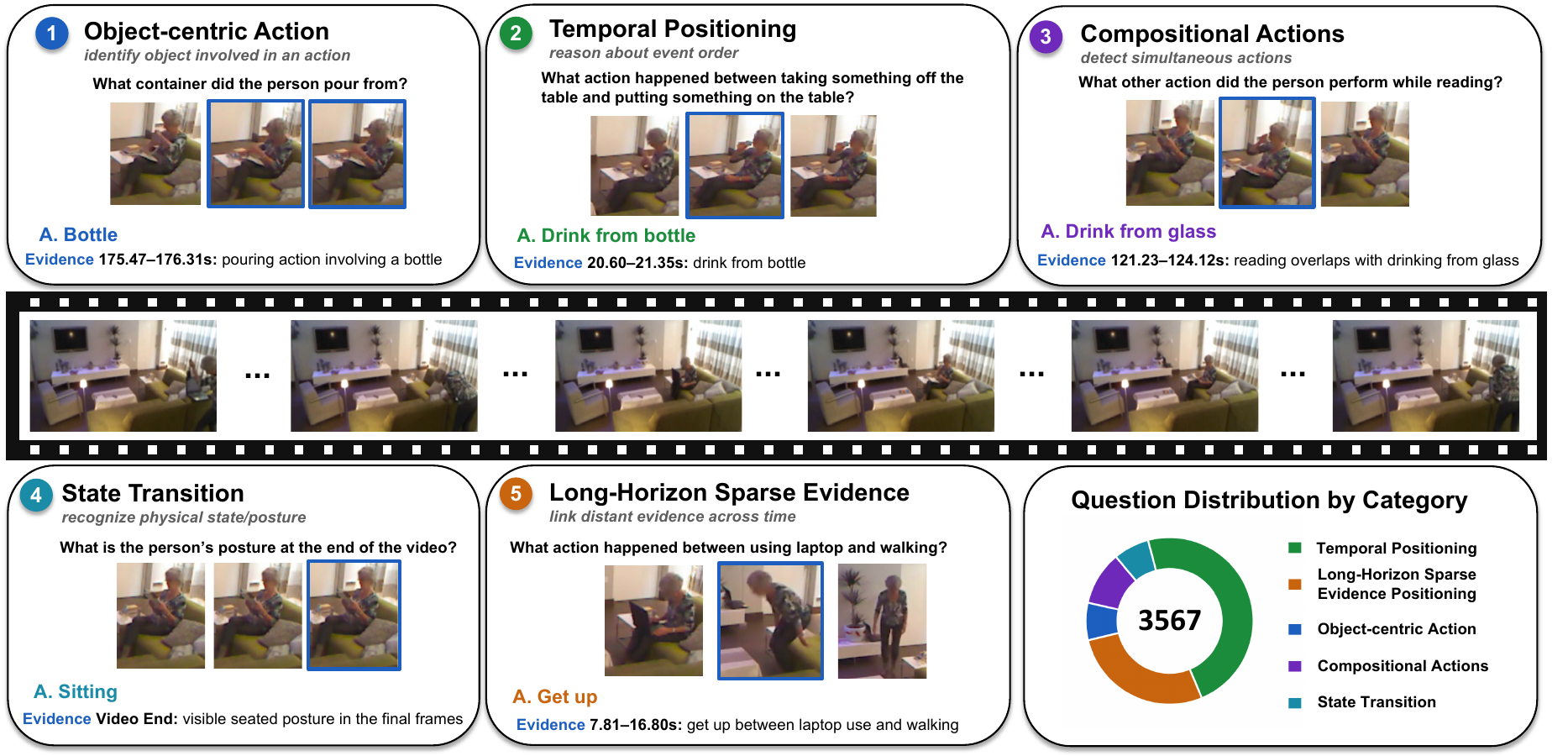}
  \caption{Overview of \benchmark~(\textsc{otb})}
  \label{fig:otb}
\end{figure*}

\section{OpenTSUBench (\textsc{otb}): Construction and Evaluation Details}
\label{app:otb}

In this section, we describe \textbf{\benchmark} (\textsc{otb}), an open-ended and temporally grounded benchmark for long-horizon Activities of Daily Living (ADL) video question answering. Existing LVQA benchmarks often evaluate models through multiple-choice questions, where answer options can implicitly guide both reasoning and temporal localization. In contrast, \textsc{otb} is designed to evaluate whether a model can answer free-form questions and identify the temporal evidence that supports its answer, as demonstrated in Table~\ref{tab:otb-vs-prior}. To this end, each question in \textsc{otb} is paired with one or more supporting video intervals, allowing us to evaluate both answer correctness and temporal grounding. Figure~\ref{fig:otb} illustrates \textsc{otb} and its different strata.

\noindent\textbf{Source Data.}
\textsc{otb} is built on Toyota Smarthome Untrimmed (TSU)~\cite{Dai_2022_PAMI}, which contains untrimmed ADL videos with dense temporal annotations. The videos cover everyday activities such as preparing drinks, cooking, eating, taking pills, cleaning, posture changes, and device use. Since these activities often overlap and vary substantially in duration, TSU provides a realistic testbed for long-video temporal reasoning. We use the timestamped action annotations as the structural basis for question generation, so every question remains linked to the action event or events that justify its answer.

\noindent\textbf{Benchmark Construction.}
We construct \textsc{otb} through a pipeline, following standard benchmarks~\cite{mvbench}, that preserves the temporal structure of TSU while producing natural open-ended questions. 

\begin{algorithm*}[t]
\caption{\modelname: Local action-guided proposal and temporal verification for LVQA.}
\label{alg:pipeline}
\scalebox{0.82}{
\begin{minipage}{1.10\linewidth}
\begin{algorithmic}[1]
\Require Video $V$, question $q$, Action Detector $f_{\mathrm{act}}$, edge LLM $f_{\mathrm{llm}}$, cloud VLM $f_{\mathrm{vlm}}$, verification budget $M$.
\Ensure Answer $a^{*}$, semantic evidence $\mathcal{S}^{*}$, visual evidence $\mathcal{V}^{*}$.
\Statex \textbf{Local Action-based Candidate Evidence (ACE)}
\State $v \gets \textsc{ExtractFeatures}(V)$
\State $\mathbf{P} \gets f_{\mathrm{act}}(v)$
       \Comment{Predict segment-level action probabilities, Eq.~\ref{eq:action_probs}}
\State $\mathcal{A} \gets \textsc{DecodeTimeline}(\mathbf{P},\theta)$
       \Comment{Construct event timeline, Eq.~\ref{eq:event_timeline}}
\State $\mathcal{W}_{\mathrm{atom}} \gets \{[s_i,e_i]:(c_i,s_i,e_i)\in\mathcal{A}\}$
\State $\mathcal{G}_q \gets f_{\mathrm{llm}}(q,\mathcal{A})$
       \Comment{Query-conditioned grouping of action windows}
\State $\mathcal{W}_{\mathrm{merge}} \gets \textsc{MergeWindows}(\mathcal{G}_q,\mathcal{A})$
       \Comment{Eq.~\ref{eq:merged_windows}}
\State $\mathcal{W}_{q} \gets \mathcal{W}_{\mathrm{atom}}\cup\mathcal{W}_{\mathrm{merge}}$
\State $\mathcal{H}_{q} \gets \textsc{GenerateHypotheses}(f_{\mathrm{llm}},q,\mathcal{A},\mathcal{W}_q)$
       \Comment{$\mathcal{H}_q=\{(a_i,w_i)\}_{i=1}^{M}$}
\State $\mathcal{H}_{q}^{*} \gets \textsc{Sort}(\mathcal{H}_{q}; R(w\mid q))$
       \Comment{Scoring and reranking, Eq.~\ref{eq:relevance_score}}
\Statex \textbf{Cloud Temporal Verification}
\For{$t=1$ \textbf{to} $\min(M,|\mathcal{H}_{q}^{*}|)$}
    \State $(a_t,w_t) \gets \mathcal{H}_{q}^{*}[t]$, \quad $w_t=[s_t,e_t]$
    \State $\widetilde{V}_t \gets V[s_t,e_t]$
          \Comment{Extract selected RGB evidence clip}
    \State $(c_t,\hat{a}_t,d_t) \gets f_{\mathrm{vlm}}(\widetilde{V}_t,q,a_t)$
          \Comment{Verify candidate answer, Eq.~\ref{eq:verified_output}}
    \If{$c_t=1$}
        \State $\mathcal{S}^{*}\gets d_t$, \quad $\mathcal{V}^{*}\gets \widetilde{V}_t$
        \State \Return $(\hat{a}_t,\mathcal{S}^{*},\mathcal{V}^{*})$
    \EndIf
\EndFor
\State \Return $\textsc{Fallback}(\mathcal{H}_{q}^{*})$
       \Comment{Return best available candidate if budget is exhausted}
\end{algorithmic}
\end{minipage}
}
\end{algorithm*}

First, we canonicalize the raw annotations into an event timeline, where each event contains an action label, start and end frame, temporal neighbors, and semantic tags such as \texttt{food}, \texttt{cleaning}, \texttt{posture}, \texttt{device}, and \texttt{container}. Overlapping actions are retained rather than collapsed, since concurrency is common in ADL videos and often necessary for answering questions. Next, we instantiate syntactic templates over the canonical timeline. Each template is tied to one stratum and parameterized by event indices, so the supporting interval is known at generation time. 

Then we subsample the generated pool to balance template type, stratum, video, target action, and answer type. An LLM is used to formalize the retained questions while preserving event references, answer semantics, and temporal support. Finally, we filter questions with ambiguous references, trivial answers, near-duplicate wording, unreliable temporal adjacency, or paraphrases that alter meaning. To validate the QA construction, two human annotators audit the benchmark for answer correctness, temporal support, and stratum assignment. 

Thus each question is paired with supporting intervals $\{[s_j,e_j]\}_{j=1}^{m_q}$. Some questions require a single localized event, while others require multiple intervals, such as verifying that one action occurs after another. These annotations are aimed to distinguish models that merely answer correctly from those that actually retrieve the relevant evidence.

%%%%%%%%%%%%%%%%%%%%%%%%%%%%%%%%%%%%%%%%%%%%%%%%%%%

\section{Algorithmic Framework of \modelname}
\label{app:algo}
Algorithm~\ref{alg:pipeline} demonstrates the algorithmic Framework of \modelname.
%%%%%%%%%%%%%%%%%%%%%%%%%%%%%%%%%%%%%%%%%%%%%%%%%%%

\section{Prompts}
\label{app:prompts}

\begin{figure}[h]
  \centering
  \includegraphics[height=5cm, width=0.5\textwidth]{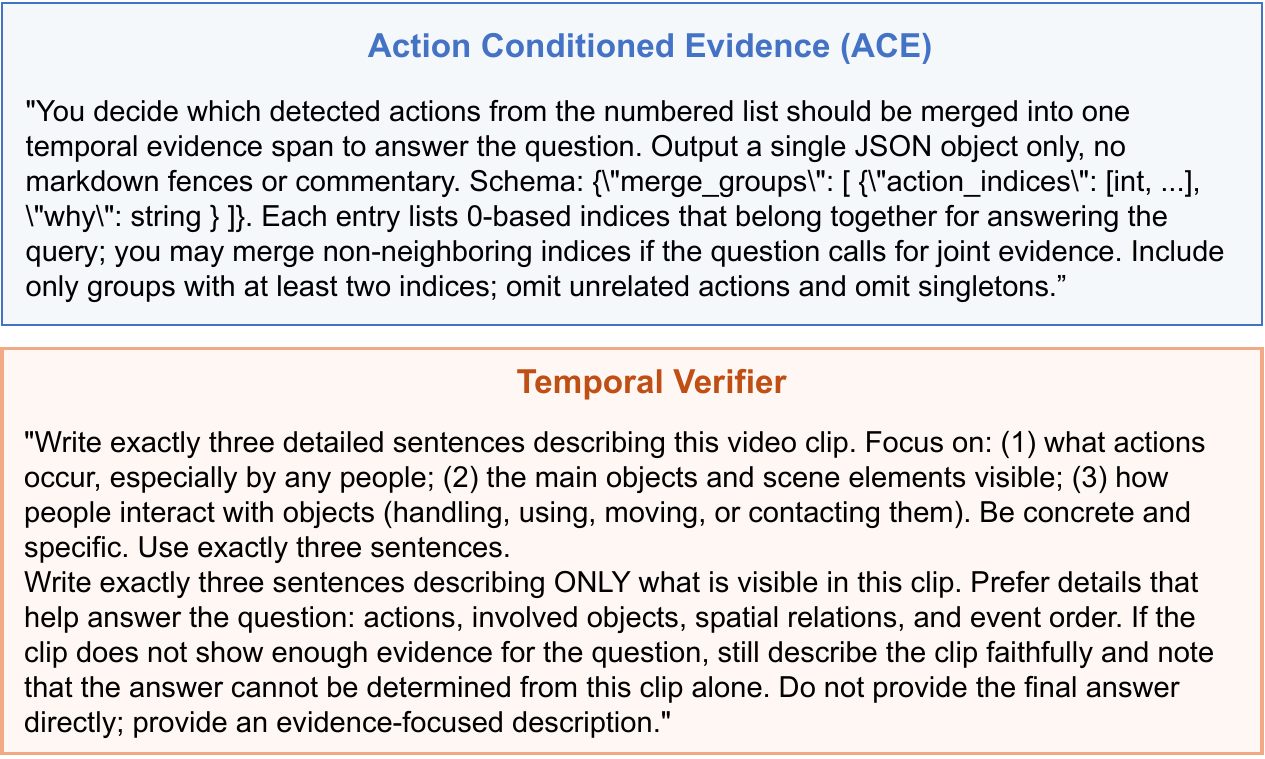}
  \caption{Prompts for ACE and Temporal Verifier in \modelname}
  \label{fig:prompt}
\end{figure}

In Figure~\ref{fig:prompt}, we provide the prompts used in ACE and the Temporal Verifier. 

\section{Qualitative Examples}
\label{app:qual}
\begin{figure*}[h]
  \centering
  \includegraphics[height=4cm, width=\textwidth]{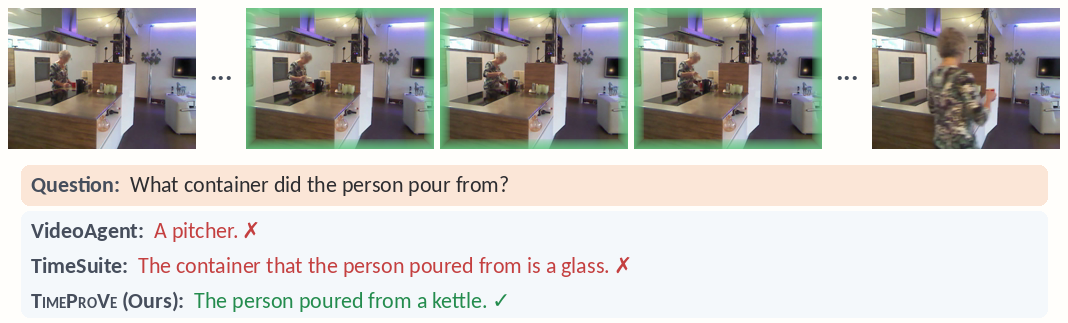}\vspace{-3mm}
  \caption{Qualitative Example for Object Centric Strata in \textsc{otb}}
  \label{fig:qual_exmpl_obj}\vspace{-2mm}
\end{figure*}

\begin{figure*}[h]
  \centering
  \includegraphics[height=4cm, width=\textwidth]{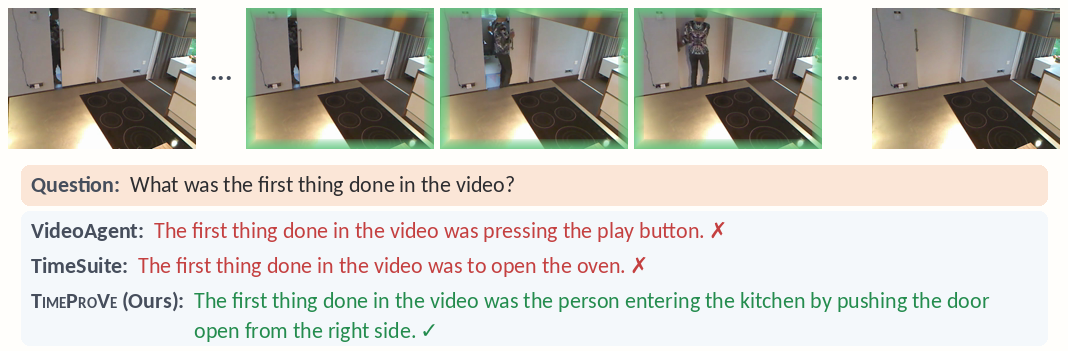}\vspace{-3mm}
  \caption{Qualitative Example for Temporal Positioning Strata in \textsc{otb}}
  \label{fig:qual_exmpl_tmp}\vspace{-2mm}
\end{figure*}

\begin{figure*}[h]
  \centering
  \includegraphics[height=4cm, width=\textwidth]{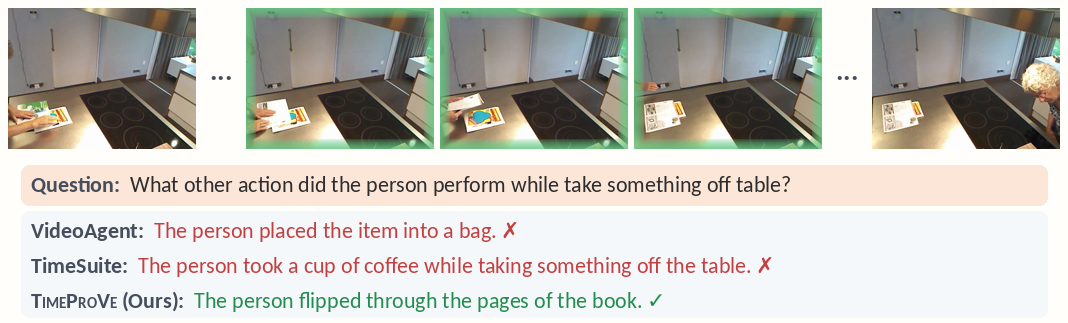}\vspace{-3mm}
  \caption{Qualitative Example for Compositional Actions Strata in \textsc{otb}}
  \label{fig:qual_exmpl_comp}\vspace{-2mm}
\end{figure*}

\begin{figure*}[h]
  \centering
  \includegraphics[height=4cm, width=\textwidth]{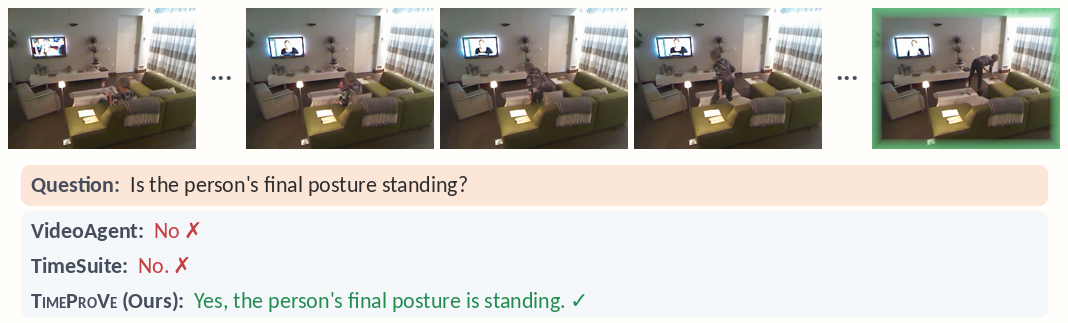}\vspace{-3mm}
  \caption{Qualitative Example for State Transition Strata in \textsc{otb}}
  \label{fig:qual_exmpl_stat}\vspace{-2mm}
\end{figure*}

\begin{figure*}[h]
  \centering
  \includegraphics[height=4cm, width=\textwidth]{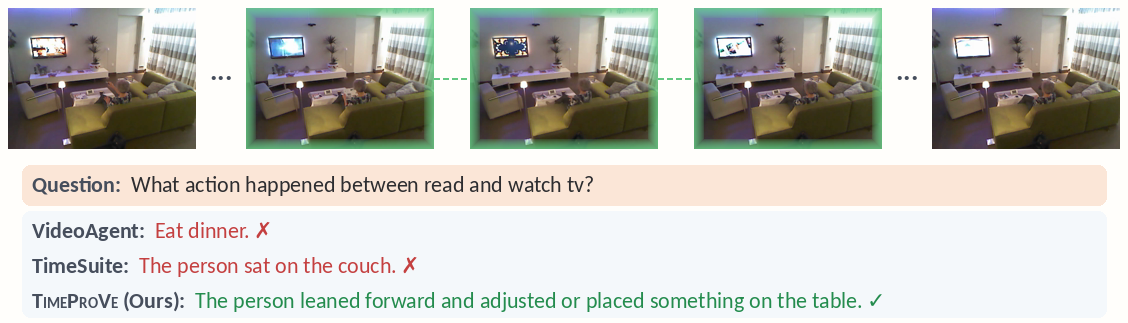}\vspace{-3mm}
  \caption{Qualitative Example for Long Horizon Sparse Evidence Strata in \textsc{otb}}
  \label{fig:qual_exmpl_lhse}\vspace{-2mm}
\end{figure*}

We demonstrate the performance of \modelname~on qualitative examples from the \textsc{otb} in Figures~\ref{fig:qual_exmpl_obj}, \ref{fig:qual_exmpl_tmp}, \ref{fig:qual_exmpl_comp}, \ref{fig:qual_exmpl_stat}, and~\ref{fig:qual_exmpl_lhse}. 


\end{document}